\pdfoutput=1
\documentclass[11pt]{article}

\usepackage[preprint]{acl}

\usepackage{times}
\usepackage{latexsym}
\usepackage[T1]{fontenc}
\usepackage[utf8]{inputenc}
\usepackage{inconsolata}
\usepackage{graphicx}
\usepackage[dvipsnames]{xcolor}

\usepackage{comment}
\usepackage{paralist}
\usepackage{booktabs}
\usepackage{shellesc}
\usepackage{amsmath}
\usepackage{tcolorbox}

\usepackage{microtype}

\title{Cascaded Batch Prompting}

\author{Sho Hoshino \and Peinan Zhang \\
CyberAgent \\
\texttt{\{hoshino\_sho,zhang\_peinan\}@cyberagent.co.jp}
}

\begin{document}
\maketitle
\begin{abstract}
Although batch prompting makes large language model inference more efficient by processing multiple instances simultaneously, it suffers from unpredictable downstream task performance.
We propose \emph{cascaded batch prompting}, a two-stage approach designed to resolve the unpredictability of conventional batch prompting by disentangling complex reasoning from symbol grounding.
Experiments on multiple-choice question answering and natural language inference demonstrate that the proposed method outperforms the standard single prompting baseline while achieving a speedup proportional to batch size, establishing a new state of the art on the Pareto frontier.
\end{abstract}

\section{Introduction}
Batch prompting is a prominent technique for making large language model (LLM) inference efficient by processing multiple instances simultaneously within a single context \cite{cheng-etal-2023-batch,lin2024batchprompt}.
However, accommodating multiple instances requires formatting modifications that can alter generated outputs.
Consequently, prior studies observe that batching can unexpectedly improve or even degrade downstream task performance, introducing a critical bottleneck of unpredictability.
This unpredictability often limits the real-world application of batch prompting at scale by forcing an undesirable choice between inference speedup and the risk of degraded performance.

\begin{figure}[t]
\centering
\includegraphics[width=\linewidth]{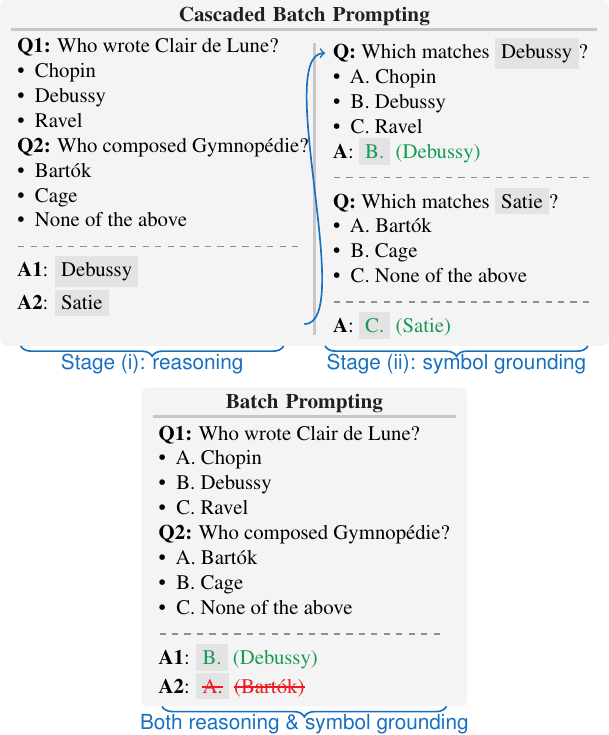}
\caption{
Cascaded batch prompting (top) disentangles reasoning from symbol grounding into a two-stage process, while conventional batch prompting (bottom) handles both in one step.
}
\label{fig:eyecatch}
\end{figure}

We hypothesize this unpredictability of conventional batch prompting is not inherent to batching itself, but it results from conflating distinct cognitive tasks into a single step.
Specifically, classification tasks impose strict output constraints that force the model to simultaneously perform:
\begin{inparaenum}[(i)]
\item the complex reasoning required to arrive at a solution, and
\item the procedural task of symbol grounding that solution in a constrained output format.
\end{inparaenum}
While the benefit of disentangling these tasks has been demonstrated for single prompting \cite{wang2024look,wang-etal-2024-answer-c}, we argue that under the pressure of batch processing, the cognitive load of this conflation becomes particularly problematic.

To test this hypothesis, we propose \emph{cascaded batch prompting}, a novel two-stage approach that disentangles reasoning from symbol grounding in the batching context (Figure~\ref{fig:eyecatch}).
Instead of forcing the LLM to generate the final constrained output directly, our method proceeds as follows:
\begin{inparaenum}[(i)]
\item a \emph{reasoning} stage that prompts the LLM to answer the question in a free-form manner, and 
\item a \emph{symbol grounding} stage that provides the LLM with its own generated answer and asks it to map that answer to the corresponding constrained output.
\end{inparaenum}
This task decomposition allows the LLM to handle each sub-task more effectively.

We evaluate our method on the diverse and challenging classification tasks of multiple-choice question answering \citep[MCQA;][]{balepur-etal-2025-best} and natural language inference \citep[NLI;][]{bowman-etal-2015-large}.
Experiments demonstrate that cascaded batch prompting outperforms the single prompting baseline while maintaining speed proportional to batch size.
These results indicate that the proposed method successfully resolves the cognitive bottleneck driving the unpredictability, establishing a new state of the art on the Pareto frontier.

\section{Prompting Methodology}
We formally define the prompting strategies for classification tasks, including MCQA and NLI.

\subsection{Problem Definition}
A classification task instance consists of an input $x$ and a set of class labels.
The desired output is the answer $a$, which indicates the correct class label.
This answer can be represented by a symbol $s$ (e.g., ``A.'') or its corresponding class label name $n$ (e.g., ``entailment'').
For any given task, there exists a deterministic function $g: s \mapsto n$ that maps a symbol to its name.
Since this mapping is one-to-one, generating either $s$ or $n$ is sufficient to determine the final answer.
We use $a$ to denote a comprehensive answer format that may include both the symbol and its name (e.g., ``A. entailment'').

We denote a call to an LLM as a function $\mathcal{G}_{X \to Y}$, which maps an input from space $X$ to an output in space $Y$.
To handle multiple instances efficiently, this function can operate on a batch of inputs, denoted by bold letters (e.g., $\mathbf{x}$).

\subsection{Single Prompting}
The standard approach is \emph{single prompting}, where the LLM is prompted to predict the answer for a single instance $x$.
Depending on the desired output format, this leads to three variants \cite{robinson2023leveraging}:
\begin{subequations}
\begin{align}
\text{End-to-End:} \quad & \hat{a} = \mathcal{G}_{x \to a}(x), \\
\text{Multiple Choice:} \quad & \hat{s} = \mathcal{G}_{x \to s}(x), \\
\text{Cloze:} \quad & \hat{n} = \mathcal{G}_{x \to n}(x).
\end{align}
\end{subequations}
In the Multiple Choice variant, the class label name can be recovered deterministically (i.e., $\hat{n} = g(\hat{s})$).
In the Cloze variant, however, mapping the free-form output back to a symbol is a non-trivial step \cite{wang2024look,wang-etal-2024-answer-c}.
While single prompting can be accurate, processing instances one by one is slow.

\subsection{Conventional Batch Prompting}
To improve throughput, \emph{conventional batch prompting} processes a batch of inputs $\mathbf{x}$ simultaneously.
Consequently, the three variants are extended accordingly:
\begin{subequations}
\begin{align}
\text{End-to-End:} \quad & \hat{\mathbf{a}} = \mathcal{G}_{x \to a}(\mathbf{x}), \\
\text{Multiple Choice:} \quad & \hat{\mathbf{s}} = \mathcal{G}_{x \to s}(\mathbf{x}), \\
\text{Cloze:} \quad & \hat{\mathbf{n}} = \mathcal{G}_{x \to n}(\mathbf{x}).
\end{align}
\end{subequations}
However, prior studies observe that its performance can be unpredictable and may suffer from degradation when batched \cite{cheng-etal-2023-batch,lin2024batchprompt}.

\subsection{Cascaded Batch Prompting}
We propose \emph{cascaded batch prompting}, a two-stage approach that disentangles the complex reasoning from the simple symbol grounding as follows.

\paragraph{Reasoning.}
First, the LLM is prompted to generate the class label name $\hat{\mathbf{n}}$ for a batch of inputs $\mathbf{x}$.
This stage allows the model to focus entirely on reasoning:
\begin{equation}
\hat{\mathbf{n}} = \mathcal{G}_{x \to n}(\mathbf{x}).
\end{equation}

\paragraph{Symbol Grounding.}
Second, the batch of generated class label names $\hat{\mathbf{n}}$ is provided as input to another LLM prompt, which maps each class label name to its corresponding symbol $\hat{\mathbf{s}}$:
\begin{equation}
\hat{\mathbf{s}} = \mathcal{G}_{n \to s}(\hat{\mathbf{n}}).\footnotemark
\end{equation}
\footnotetext
{
We also supply $\mathbf{x}$ as additional context to improve performance, though we confirmed it is not strictly necessary.
For this symbol grounding stage, we processed each instance individually via single prompting; we found this approach robust for this procedural task, and the computational overhead was minimal in our experiments.
}

The complete cascaded process is a composition of these two steps, expressed as:
\begin{equation}
\hat{\mathbf{s}} = \mathcal{G}_{n \to s}(\mathcal{G}_{x \to n}(\mathbf{x})).
\end{equation}
By separating the generation process, we hypothesize that the LLM can handle each sub-task more effectively, thereby overcoming the limitations of conventional batch prompting.

\section{Experiments}
We evaluate various prompting strategies using the challenging tasks of MCQA and NLI, configured as follows.\footnote
{We provide full implementation details, including hyperparameters and data splitting, in Appendix~\ref{appendix:detail}.}

\subsection{Setup}

\paragraph{Data.}
We use the Massive Multitask Language Understanding \citep[MMLU;][]{hendrycks2021measuring} dataset for MCQA and the Multi-Genre Natural Language Inference \citep[MNLI;][]{williams-etal-2018-broad} dataset for NLI.
We selected these benchmarks because they are commonly used in popular LLM benchmarks, including BIG-bench \cite{srivastava2023beyond} and HELM \cite{liang2023holistic}.

\paragraph{Metrics.}
We report classification accuracy, the standard metric used for MCQA and NLI.

\paragraph{LLMs.}
We primarily use GPT-4.1 and its smaller variant GPT-4.1-mini \cite{openai2024gpt4o} with batch sizes varying from 1 to 128 to assess performance robustness across model and batch scales.
We additionally use open-weight Phi-4 \cite{abdin2024phi4technicalreport}.
We employ nucleus sampling \cite{holtzman2020curious} with top-$p$=0.9, without changing other hyperparameters unless explicitly mentioned.

\begin{table}[t]
\centering
\footnotesize
\begin{tabular}{@{}c@{}ccc@{}}
\toprule
& & \multicolumn{2}{c}{Accuracy (\%) $\uparrow$} \\
\cmidrule(l){3-4}
Prompt & Batch Size & MMLU & MNLI-m \\
\midrule
\multicolumn{4}{c}{\textsc{GPT-4.1}} \\
Single & -- & 84.39 & 82.19 \\
Batch & 32 & 85.51 & \textbf{86.14}\rlap{$^*$} \\
Cascaded Batch & 32 & \textbf{86.81}\rlap{$^*$} & 85.09 \\
\midrule
\multicolumn{4}{c}{\textsc{GPT-4.1-mini}} \\
Single & -- & 80.12 & 82.60 \\
Batch & 32 & 81.88 & 85.60 \\
Cascaded Batch & 32 & \textbf{82.96}\rlap{$^*$} & \textbf{86.20}\rlap{$^*$} \\
\midrule
\multicolumn{4}{c}{\textsc{Phi-4}} \\
Single & -- & 74.34 & 82.15 \\
Batch & 32 & 58.35 & 83.19 \\
Cascaded Batch & 32 & \textbf{77.18}\rlap{$^*$} & \textbf{83.57} \\
\bottomrule
\end{tabular}
\caption{
Comparison of various prompting strategies.
For a fair comparison, we chose the batch size of 32 that performed best for the conventional batch prompting baseline.
Bold text highlights the highest scores.
The asterisk ($^*$) denotes statistical significance ($p < 0.05$, detailed in Appendix~\ref{appendix:detail}).
}
\label{tab:main}
\end{table}

\begin{figure}[t]
\centering
\includegraphics[width=\linewidth]{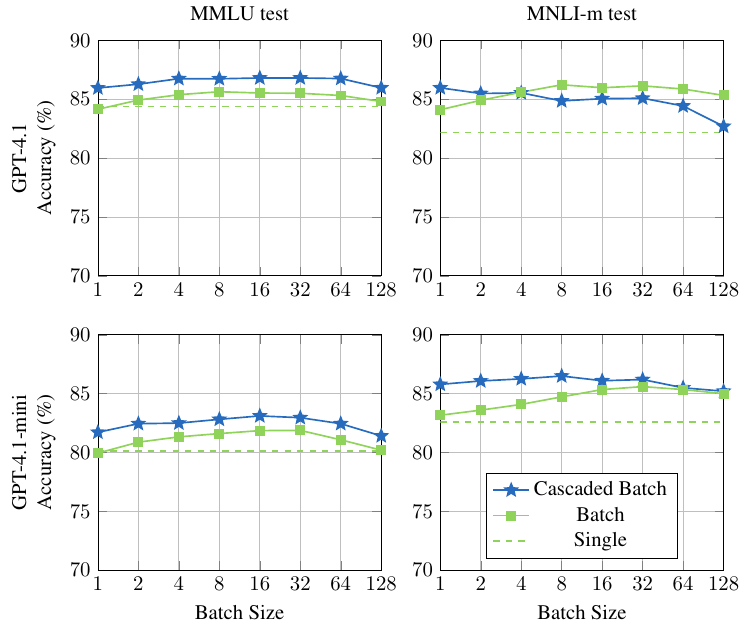}
\caption{
Accuracy scaling with batch size.
Cascaded batch prompting demonstrates robust performance across tasks and models.
Its advantage is most pronounced with GPT-4.1-mini (bottom row), where it consistently outperforms the baselines.
}
\label{fig:scaling}
\end{figure}

\subsection{Main Results}
\label{sec:main}
Table~\ref{tab:main} presents our main results.
Across all three models, our cascaded batch prompting achieves the highest accuracy on both MMLU (86.81\%) and MNLI (86.20\%).
With GPT-4.1, cascaded batch prompting obtains the top score on MMLU.
On MNLI, however, conventional batch prompting performs slightly better.
We attribute this to the fact that the NLI task inherently lacks a free-form output space (i.e., it only requires a predefined symbolic word such as ``entailment''), making the additional symbol grounding step less critical for a more powerful model.
Nevertheless, the advantage of our approach is underscored with GPT-4.1-mini and Phi-4, where cascaded batch prompting is the best performer on both tasks, which suggests our method is most beneficial when models face higher cognitive loads.
These results support our core claim that disentangling reasoning from symbol grounding is a robust strategy that consistently improves upon single prompting.

\subsection{Scalability Analysis}
\label{sec:scaling}
Figure~\ref{fig:scaling} shows that cascaded batch prompting's accuracy remains robust as batch size increases.
In contrast, conventional batch prompting's performance is less consistent, and can underperform the single prompting baseline (e.g., with Phi-4 on MMLU).
This result confirms that our method effectively leverages batching's efficiency without sacrificing accuracy.

A closer inspection of Figure~\ref{fig:scaling} reveals a slight degradation in performance for both batch methods at the largest batch size of 128.
We attribute this not to a decline in reasoning capability but rather to a mechanical artifact of processing batches, where the model may occasionally fail to produce a corresponding output for every input instance.
This can result in a mismatch between the number of input and output lines.
A simple sanity check can identify and rerun the affected instances to rectify these issues (detailed in Section~\ref{sec:sanity}).
For the clarity of our analysis, however, we did not include these post hoc corrections in our main results.

\begin{figure}[t]
\centering
\includegraphics[width=\linewidth]{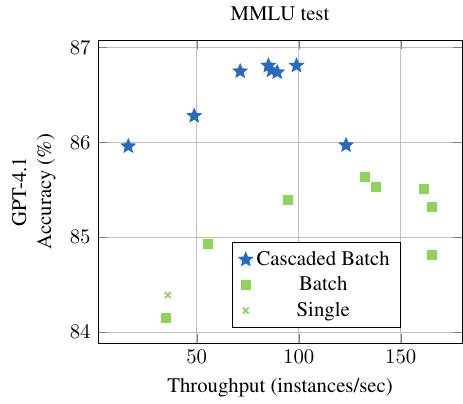}
\caption{
Compared with single prompting, cascaded batch prompting achieves higher accuracy and throughput at the same time, establishing a new Pareto frontier. 
Meanwhile, conventional batch prompting underperforms, demonstrating its unpredictability.
}
\label{fig:runtime}
\end{figure}

\subsection{Runtime Analysis}
\label{sec:runtime}
Figure~\ref{fig:runtime} visualizes the trade-off between accuracy and instance-level throughput (instances/sec).
While single prompting offers high accuracy at low speed, conventional batch prompting excels at speed while suffering from unpredictable performance, even underperforming the single prompting baseline.
Our cascaded approach resolves this unpredictability, establishing a new Pareto frontier by achieving higher accuracy and throughput at the same time.

\subsection{Ablation Study}
\begin{figure}[t]
\centering
\includegraphics[width=\linewidth]{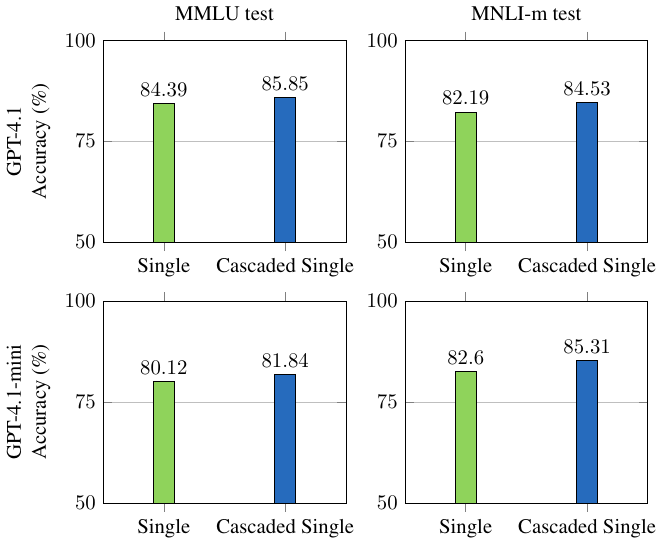}
\caption{
Cascaded single prompting outperforms the conventional baseline, demonstrating the effectiveness of the two-stage design even without batching.
}
\label{fig:ablation}
\end{figure}

To isolate the effect of our two-stage design from the effects of batching, we conducted an ablation study comparing  standard single prompting with cascaded single prompting \cite{wang2024look,wang-etal-2024-answer-c}.
As shown in Figure~\ref{fig:ablation}, cascaded single prompting consistently outperforms its standard counterpart across both MMLU and MNLI.
This result suggests that the performance benefit is not solely due to batching but is fundamentally rooted in the task decomposition itself.

\subsection{Sanity Check}
\label{sec:sanity}
As discussed in Section~\ref{sec:scaling}, we can also add a simple sanity check during batch prompting inference to verify if the number of outputs matches the number of inputs.
Implementing this sanity check prevents a single missing generation from misaligning all subsequent answers in the batch.
This sanity check can be applied to both conventional and cascaded batch prompting and is therefore orthogonal.

Table~\ref{tab:sanity} reports preliminary results using GPT-4.1-mini and the MMLU test set.
While this optional and simple step cannot fully resolve the performance decline as batch size increases, it successfully mitigates input and output misalignment.
However, the resulting performance improvement is slight and not statistically significant.

\subsection{Cost Analysis}
\label{sec:cost}
Table~\ref{tab:cost} reports the inference cost in terms of token-level throughput (tokens/sec) on the MMLU test set using GPT-4.1, distinct from the instance-level throughput discussed in Section~\ref{sec:runtime}.
Since cascaded batch prompting requires an additional inference stage, it processes more tokens overall, resulting in a higher token-level throughput and a cost approximately 1.2 times higher than that of conventional batch prompting.
We emphasize that this modest overhead is a necessary trade-off to resolve the unpredictability of conventional batch prompting.

\begin{table}[t]
\centering
\footnotesize
\begin{tabular}{@{}l@{}cc@{}}
\toprule
& & Accuracy (\%) $\uparrow$ \\
\cmidrule(l){3-3}
\multicolumn{1}{c}{Prompt} & Batch Size & MMLU \\
\midrule
\multicolumn{3}{c}{\textsc{GPT-4.1-mini}} \\
Cascaded Batch \\
\quad w/o Sanity Check & 32 & \textbf{82.96} \\
& 128 & 81.41 \\
\midrule
\quad w/ Sanity Check & 128 & 81.61 \\
\bottomrule
\end{tabular}
\caption{
Impact of the sanity check on cascaded batch prompting.
The sanity check partially recovers performance lost to input and output misalignment, although it cannot fully resolve the performance decline at larger batch sizes.
}
\label{tab:sanity}
\end{table}

\begin{table}[t]
\centering
\footnotesize
\begin{tabular}{@{}c@{}ccc@{}}
\toprule
& & \multicolumn{2}{c}{Throughput (tokens/sec) $\downarrow$} \\
\cmidrule(l){3-4}
& & \multicolumn{2}{c}{MMLU} \\
\cmidrule(l){3-4}
Prompt & Batch Size & Input & Output \\
\midrule
\multicolumn{4}{c}{\textsc{GPT-4.1}} \\
Single & -- & 4,910 & 47 \\
Batch & 32 & 18,990 & 965 \\
Cascaded Batch & 32 & 23,341 & 1,220 \\
\bottomrule
\end{tabular}
\caption{
Cascaded batch prompting introduces a modest cost overhead due to the additional symbol grounding stage.
}
\label{tab:cost}
\end{table}

\section{Discussion}
Our experiments have demonstrated the empirical advantages of cascaded batch prompting.
We now turn to a discussion of the underlying mechanisms that drive its effectiveness and the broader implications of task decomposition.

\paragraph{Cognitive Division of Labor.}
We attribute the effectiveness of our method to a cognitive division of labor, analogous to Adam Smith's ``pin factory'' \cite{smith1776wealth}.
Our cascaded approach separates the process into a reasoning stage for problem-solving and a symbol grounding stage for procedural formatting.

\paragraph{Pragmatic Value of Task Decomposition.}
While developing end-to-end LLMs is a prominent research goal, our results show the pragmatic value of task decomposition.
Since current models can struggle with the combined cognitive load of reasoning and symbol grounding under batch pressure, principled decomposition remains a critical and effective strategy.

\section{Conclusion}
In this study, we identified that the unpredictability of conventional batch prompting results from conflating distinct cognitive tasks into a single step.
To address this, we proposed cascaded batch prompting, a two-stage approach that disentangles complex reasoning from simple symbol grounding, thereby resolving this unpredictability.
Experiments demonstrate that the proposed method outperforms the single prompting baseline while achieving a speedup proportional to batch size.
While our experiments focused on multiple-choice question answering and natural language inference, the core principle extends to a wide range of classification tasks where a model's free-form reasoning must be mapped to a constrained output format.

\section*{Limitations}

\paragraph{Task Applicability.}
Our evaluation intentionally focuses on classification tasks to provide a controlled setting for validating the disentanglement of reasoning from symbol grounding.
To extend this two-stage approach to the challenges of open-ended generation, future work can investigate integrating methods like universal self-consistency \cite{chen2024universal}.

\paragraph{Inference Overhead.}
The proposed cascaded batch prompting uses a two-stage inference process that adds overhead compared to conventional single-stage prompting (detailed in Section~\ref{sec:cost}).
For a dataset of size $N$ and a batch size of $b$, conventional batch prompting requires $N/b$ inference calls.
In contrast, our implementation requires $N/b$ calls for the batched reasoning stage plus an additional $N$ individual calls for the unbatched symbol grounding stage.
We consider this overhead a necessary trade-off to resolve the unpredictability of conventional batch prompting and therefore worthwhile.

\bibliography{custom}

\clearpage
\appendix

\section{Setup Details}
\label{appendix:detail}

\paragraph{Data Splitting.}
For the MMLU dataset, we use the validation split for development and the test split for testing.
For the MNLI dataset, similarly, we use the matched validation split for development and the matched test split for testing.

\paragraph{Hyperparameters.}
We set max tokens to 20 for single prompting and 1,000 for other prompting strategies.

\paragraph{Model Versions.}
We use GPT-4.1 version 2025-04-14 and GPT-4.1-mini version 2025-04-14 \cite{openai2024gpt4o} with Phi-4 version 7 \cite{abdin2024phi4technicalreport}.
To maximize throughput, all calls to the LLM API were made asynchronously.

\paragraph{Prompts.}
\label{appendix:prompt}
Figures~\ref{fig:prompt-mcqa}, \ref{fig:prompt-nli}, and \ref{fig:prompt-batch} list our prompts used for the tasks of MCQA and NLI.

\paragraph{Statistical Testing.}
We assessed statistical significance using paired bootstrap resampling \cite{koehn-2004-statistical}.
The performance gains from cascaded batch prompting are statistically significant ($p < 0.05$) on the MMLU benchmark across the evaluated models.
The performance gains on MNLI were mostly consistent in direction, but did not reach statistical significance with Phi-4 and underperformed the conventional batch prompting baseline with GPT-4.1.
This suggests limited gains on the NLI task as discussed in Section~\ref{sec:main}.

\paragraph{Licenses.}
We performed experiments using the MMLU dataset \cite{hendrycks2021measuring} released under the MIT license\footnote{ \url{https://huggingface.co/datasets/cais/mmlu}}, the MNLI dataset \cite{williams-etal-2018-broad} available under a collection of licenses\footnote
{
The majority of the corpus is released under the OANC's license.
Data in the fiction section falls under several permissive licenses, including Creative Commons Share-Alike 3.0 Unported, Creative Commons Attribution 3.0 Unported, and the public domain in the United States.
}
\footnote{\url{https://huggingface.co/datasets/nyu-mll/multi_nli}}, GPT-4.1 API \cite{openai2024gpt4o} released under a proprietary license via Azure OpenAI Service\footnote{\url{https://azure.microsoft.com/en-us/products/ai-services/openai-service}}, and Phi-4 \cite{abdin2024phi4technicalreport} released under the MIT license\footnote{\url{https://huggingface.co/microsoft/phi-4}}.

\begin{figure}[t]
\centering
\footnotesize
\begin{tcolorbox}[boxrule=0.5pt, top=0pt, bottom=0pt, left=0pt, right=0pt, colback=NavyBlue!25, title=Stage (i): Reasoning]
Please read the multiple-choice question below carefully and select ONE of the listed options. Provide only the answers. Keep them as short as possible. Do not output anything else after that. \\
Note: The number of answers should be the same as the number of the questions. \\
\\
Question[1]: \{question\} \\
Options[1]: \\
\{options\} \\
Question[2]: \{question\} \\
Options[2]: \\
\{options\} \\
... \\
\\
Example: \\
Answer[1]: X. \\
Answer[2]: X. \\
... \\
\\
Answer:
\end{tcolorbox}
\begin{tcolorbox}[boxrule=0.5pt, top=0pt, bottom=0pt, left=0pt, right=0pt, colback=NavyBlue!25, title=Stage (ii): Symbol Grounding]
Given a question, a list of options, and an answer, select the one option that matches the answer. Provide only the symbols and do not output anything else after that. \\
\\
Question: \{question\} \\
Options: \\
\{options\} \\
Potential Answer: \{answer\} \\
\\
Answer: 
\end{tcolorbox}
\caption{
Listing cascaded batch prompts used in the MCQA task.
The placeholders ``\{question\}'', ``\{options\}'', and ``\{answer\}'' are replaced with the actual question, its options, and LLM answer.
}
\label{fig:prompt-mcqa}
\end{figure}

\begin{figure}[t]
\centering
\footnotesize
\begin{tcolorbox}[boxrule=0.5pt, top=0pt, bottom=0pt, left=0pt, right=0pt, colback=NavyBlue!25, title=Stage (i): Reasoning]
Given a premise and a hypothesis, what is the semantic relation between them? Provide only the answers. Keep them as short as possible. Do not output anything else after that. \\
Note: The number of answers should be the same as the number of the questions. \\
\\
Premise[1]: \{premise\} \\
Hypothesis[1]: \{hypothesis\} \\
Options[1]: \\
- entailment \\
- neutral \\
- contradiction \\
Premise[2]: \{premise\} \\
Hypothesis[2]: \{hypothesis\} \\
Options[2]: \\
... \\
\\
Example: \\
Answer[1]: X. \\
Answer[2]: X. \\
... \\
\\
Answer:
\end{tcolorbox}
\begin{tcolorbox}[boxrule=0.5pt, top=0pt, bottom=0pt, left=0pt, right=0pt, colback=NavyBlue!25, title=Stage (ii): Symbol Grounding]
Given a question, a list of options, and an answer, select the one option that matches the answer. Provide only the symbols and do not output anything else after that. \\
\\
Question: Given a premise and a hypothesis, what is the semantic relation between them? \\
Premise[1]: \{premise\} \\
Hypothesis[1]: \{hypothesis\} \\
Options: \\
A. entailment \\
B. neutral \\
C. contradiction \\
Potential Answer: \{answer\} \\
\\
Answer: 
\end{tcolorbox}
\caption{
Listing cascaded batch prompts used in the NLI task.
The placeholders ``\{premise\}'', ``\{hypothesis\}'', and ``\{answer\}'' are replaced with the actual premise, hypothesis, and LLM answer.
}
\label{fig:prompt-nli}
\end{figure}

\begin{figure}[t]
\centering
\footnotesize
\begin{tcolorbox}[boxrule=0.5pt, top=0pt, bottom=0pt, left=0pt, right=0pt, colback=YellowGreen!25, title=Conventional Batch Prompting for MCQA]
Please read the multiple-choice question below carefully and select ONE of the listed options. Provide only the symbols and do not output anything else after that. \\
Note: The number of answers should be the same as the number of the questions. \\
\\
Question[1]: \{question\} \\
Options[1]: \\
\{options\} \\
Question[2]: \{question\} \\
Options[2]: \\
\{options\} \\
... \\
\\
Example: \\
Answer[1]: X. \\
Answer[2]: X. \\
... \\
\\
Answer:
\end{tcolorbox}
\begin{tcolorbox}[boxrule=0.5pt, top=0pt, bottom=0pt, left=0pt, right=0pt, colback=YellowGreen!25, title=Conventional Batch Prompting for NLI]
Given a premise and a hypothesis, what is the semantic relation between them? Provide only the symbols and do not output anything else after that. \\
Note: The number of answers should be the same as the number of the questions. \\
\\
Premise[1]: \{premise\} \\
Hypothesis[1]: \{hypothesis\} \\
Options[1]: \\
A. entailment \\
B. neutral \\
C. contradiction \\
Premise[2]: \{premise\} \\
Hypothesis[2]: \{hypothesis\} \\
Options[2]: \\
... \\
\\
Example: \\
Answer[1]: X. \\
Answer[2]: X. \\
... \\
\\
Answer:
\end{tcolorbox}
\caption{
Listing conventional batch prompts used in the MCQA and NLI tasks, respectively.
Unlike cascaded batch, these prompts are tailored with symbol grounding in mind.
}
\label{fig:prompt-batch}
\end{figure}

\end{document}